\def\year{2017}\relax
\documentclass[letterpaper]{article} 
\usepackage{aaai17}  
\usepackage{times}  
\usepackage{helvet}  
\usepackage{courier}  
\usepackage{url}  
\usepackage{graphicx}  
\usepackage{subfiles}
\usepackage[table]{xcolor}

\usepackage{algorithm}
\usepackage{algpseudocode}

\title{Combining Strategic Learning and Tactical Search in Real-Time Strategy Games}

\author{Nicolas~A.~Barriga \and Marius~Stanescu \and Michael~Buro\\
Department of Computing Science\\
University of Alberta, Canada\\
\{barriga\textbar astanesc\textbar mburo\}@ualberta.ca
}

\begin{document}

\maketitle

\begin{abstract}
A commonly used technique for managing AI complexity in real-time strategy (RTS) games is to 
use action and/or state abstractions. High-level abstractions can often lead to good strategic 
decision making, but tactical decision quality may suffer due to lost details. A competing method is to sample the search space which often leads to good tactical performance in simple
scenarios, but poor high-level planning. 

We propose to use a deep convolutional neural network~(CNN) to select among a
limited set of abstract action choices, and to utilize the remaining
computation time for game tree search to improve low level tactics. The CNN is
trained by supervised learning on game states labelled by \emph{Puppet Search},
a strategic search algorithm that uses action abstractions. The network is
then used to select a script --- an abstract action --- to produce low level
actions for all units. Subsequently, the game tree search algorithm improves
the tactical actions of a subset of units using a limited view of the game
state only considering units close to opponent units.

Experiments in the $\mu$RTS game show that the combined algorithm results in
higher win-rates than either of its two independent components and other
state-of-the-art $\mu$RTS agents.

To the best of our knowledge, this is the first successful application of a 
convolutional network to play a full RTS game on standard game maps, as previous
work has focused on sub-problems, such as combat, or on very small maps.

\end{abstract}

\section{Introduction}\label{sec:intro}

In recent years, numerous challenging research problems have attracted AI
researchers to using real-time strategy~(RTS) games as test-bed in several
areas, such as case-based reasoning and planning~\cite{ontanon2007cbr},
evolutionary computation~\cite{barriga2014building}, machine
learning~\cite{synnaeve2011aiide}, deep
learning~\cite{usunier2016episodic,Foerster2017,Peng2017} and heuristic and
adversarial search~\cite{churchill2011build,barriga2015}. Functioning AI
solutions to most RTS sub-problems exist, but combining those doesn't come
close to human level performance\footnote{\url{http://www.cs.mun.ca/~dchurchill/starcraftaicomp/report2015.shtml#mvm}}.

To cope with large state spaces and branching factors in RTS games, recent
work focuses on smart sampling of the search space
\cite{churchill2013portfolio,ontanon2017combinatorial,ontanon2016informed,ontanon2015adversarial}
and state and action abstractions
\cite{uriarte2014game,stanescu2014,barriga2017game}.  The first approach
produces strong agents for small scenarios. The latter techniques work well on
larger problems because of their ability to make good strategic
choices. However, they have limited tactical ability, due to their necessarily
coarse-grained abstractions. One compromise would be to allocate computational
time for search-based approaches to improve the tactical decisions, but this
allocation would come at the expense of allocating less time to strategic
choices.

We propose to train a deep convolutional neural network~(CNN) to predict the
output of \emph{Puppet Search}, thus leaving most of the time free for use by
a tactical search algorithm. \emph{Puppet Search} is a strategic search
algorithm that uses action abstractions and has shown good results,
particularly in large scenarios. We will base our network on previous work on
CNNs for state evaluation~\cite{stanescu2016evaluating}, reformulating the
earlier approach to handle larger maps.

This paper's contributions are a network architecture capable of scaling to larger
map sizes than previous approaches, a policy network for selecting high-level
actions, and a method of combining the policy network with a tactical search
algorithm that surpasses the performance of both individually.

The remainder of this paper is organized as follows: Section~\ref{sec:background} 
discussed previous related work, Section~\ref{sec:implementation} describes our 
proposed approach and Section~\ref{sec:results} provides experimental results. 
We then conclude and outline future work.

\begin{table*}[ht]
\begin{center}
{\caption{Input feature planes for Neural Network. 25 planes for the evaluation 
network and 26 for the policy network.}\label{table:features}}\medskip
\begin{tabular}{|l|c|l|}
\hline
Feature				&\# of planes	&Description	\\
\hline
Unit type 			&6 			&Base, Barracks, worker, light, ranged, heavy \\
Unit health points 	&5 			&$1, 2, 3, 4, \mbox{or} \ge 5$ \\
Unit owner 			&2 			&Masks to indicate all units belonging to one player  \\
Frames to completion &5 		&$0\!-\!25,\, 26\!-\!50,\, 51\!-\!80,\, 81\!-\!120,\, \mbox{or} \ge 121$ \\
Resources 			&7 			&$1,\, 2,\, 3,\, 4,\, 5,\, 6\!-\!9,\, \mbox{or} \ge 10$ \\
\hline
Player              &1          &Player for which to select strategy\\
\hline
\end{tabular}
\end{center}
\end{table*}

\section{Related Work}\label{sec:background}

Ever since the revolutionary results in the \emph{ImageNet}
competition~\cite{krizhevsky2012imagenet}, CNNs have been applied successfully
in a wide range of domains. Their ability to learn hierarchical structures of
spatially invariant local features make them ideal in settings that can be
represented spatially. These include uni-dimensional streams in natural
language processing~\cite{collobert2008unified}, two-dimensional board
games~\cite{silver2016mastering}, or three-dimensional video
analysis~\cite{ji20133d}.

These diverse successes have inspired the application of CNNs to games. They 
have achieved human-level performance in several \emph{Atari} games, by using 
Q-learning, a well known reinforcement learning~(RL) 
algorithm~\cite{mnih2015human}. But the most remarkable accomplishment may be 
AlphaGo~\cite{silver2016mastering}, a \emph{Go} playing program that last year 
defeated Lee Sedol, one of the top human professionals, a feat that was thought 
to be at least a decade away. As much an engineering as a scientific 
accomplishment, it was achieved using a combination of tree search and a series 
of neural networks trained on millions of human games and self-play, running on 
thousands of CPUs and hundreds of GPUs.

These results have sparked interest in applying deep learning to games with
larger state and action spaces. Some limited success has been found in
micromanagement tasks for RTS games~\cite{usunier2016episodic}, where a deep
network managed to slightly outperform a set of baseline
heuristics. Additional encouraging results were achieved for the task of
evaluating RTS game states~\cite{stanescu2016evaluating}. The network
significantly outperforms other state-of-the-art approaches at predicting game
outcomes. When it is used in adversarial search algorithms, they perform
significantly better than using simpler evaluation functions that are three to
four orders of magnitude faster.

Most of the described research on deep learning in multi-agent domains assumes
full visibility of the environment and lacks communication between
agents. Recent work addresses this problem by learning communication between
agents alongside their policy \cite{Sukhbaatar2016}. In their model, each
agent is controlled by a deep network which has access to a communication
channel through which they receive the summed transmissions of other
agents. The resulting model outperforms models without communication,
fully-connected models, and models using discrete communication on simple
imperfect information combat tasks.  However, symmetric communication prevents
handling heterogeneous agent types, limitation later removed by
\cite{Peng2017} which use a dedicated bi-direction communication channel and
recurrent neural networks. This would be an alternative to the search
algorithm we use for the tactical module on
section~\ref{sec:strategyandtactics}, in cases where there is no forward model
of the game, or there is imperfect information.

A new search algorithm that has shown good results particularly in large RTS 
scenarios, is \emph{Puppet 
Search}~\cite{barriga2015,barriga2017combining,barriga2017game}. It is an action 
abstraction mechanism that uses fast scripts with a few carefully selected 
choice points. These scripts are usually hard-coded strategies, and the number of choice
points will depend on the time constraints the system has to meet. 
These choice points are then exposed to an adversarial look-ahead 
procedure, such as Alpha-Beta or Monte Carlo Tree Search~(MCTS). The algorithm 
then uses a forward model of the game to examine the outcome of different 
choice combinations and decide on the best course of action. Using a restricted 
set of high-level actions results in low branching factor, enabling deep 
look-ahead and favouring strong strategic decisions. Its main weakness is its 
rigid scripted tactical micromanagement, which led to modest results on small 
sized scenarios where good micromanagement is key to victory. 

\section{Algorithm Details}\label{sec:implementation}

\begin{figure}[t]
\centerline{\includegraphics[width=\columnwidth]{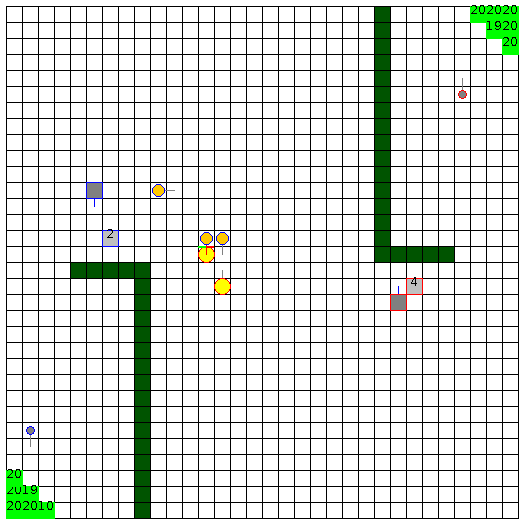}}
\caption{$\mu$RTS screenshot from a match between scripted LightRush and
  HeavyRush agents. Light green squares are resources, dark green are walls,
  dark grey are barracks and light grey the bases. Numbers indicate
  resources. Grey circles are worker units, small yellow circles are light
  combat units and big yellow ones are heavy combat units. Blue lines show
  production, red lines an attack and grey lines moving direction. Units are
  outlined blue (player 1) and red (player 2). $\mu$RTS can be found at
  {\small\url{https://github.com/santiontanon/microrts}}.} \label{fig:microRTS}
\end{figure}

We build on previous work on RTS game state
evaluation~\cite{stanescu2016evaluating} applied to $\mu$RTS~(see
figure~\ref{fig:microRTS}). This study presented a neural network architecture
and experiments comparing it to simpler but faster evaluation functions.  The
CNN-based evaluation showed a higher accuracy at evaluating game states. In
addition, when used by state-of-the-art search algorithms, they perform
significantly better than the faster evaluations.  Table~\ref{table:features}
lists the input features their network uses.

The network itself is composed of two convolutional layers followed by two
fully connected layers. It performed very well on 8$\times$8 maps. However, as
the map size increases, so does the number of weights on the fully connected
layers, which eventually dominates the weight set. To tackle this problem, we
designed a fully convolutional network~(FCN) which only consists of
intermediate convolutional layers~\cite{springenberg2014striving} and has the
advantage of being an architecture that can fit a wide range of board sizes.

\begin{table}[t]
  \centering
  \vspace{-1.5em}
  \caption{Neural Network Architecture}\medskip
  \label{table:network}
  \begin{tabular}{|c|c|}
  \hline
  Evaluation Network& Policy Network \\
  \hline
  \hline
  Input 128x128, 25 planes    &Input 128x128, 26 planes    \\
  \hline
  \multicolumn{2}{|c|}{2x2 conv. 32 filters, pad 1, stride 1, LReLU}\\
  \multicolumn{2}{|c|}{Dropout 0.2}\\
  \hline
  \multicolumn{2}{|c|}{3x3 conv. 32 filters, pad 0, stride 2, LReLU}\\
  \multicolumn{2}{|c|}{Dropout 0.2}\\
  \hline
  \multicolumn{2}{|c|}{2x2 conv. 48 filters, pad 1, stride 1, LReLU}\\
  \multicolumn{2}{|c|}{Dropout 0.2}\\
  \hline
  \multicolumn{2}{|c|}{3x3 conv. 48 filters, pad 0, stride 2, LReLU}\\
  \multicolumn{2}{|c|}{Dropout 0.2}\\
  \hline
  \multicolumn{2}{|c|}{2x2 conv. 64 filters, pad 1, stride 1, LReLU}\\
  \multicolumn{2}{|c|}{Dropout 0.2}\\
  \hline
  \multicolumn{2}{|c|}{3x3 conv. 64 filters, pad 0, stride 2, LReLU}\\
  \multicolumn{2}{|c|}{Dropout 0.2}\\
  \hline
  \multicolumn{2}{|c|}{1x1 conv. 64 filters, pad 0, stride 1, LReLU}\\
   \hline
  1x1 conv. 2 filters     &1x1 conv. 4 filters\\
   pad 0, stride 1, LReLU & pad 0, stride 1, LReLU\\
  \hline 
  \multicolumn{2}{|c|}{Global averaging over 16x16 planes}\\
  \hline
  2-way softmax     &4-way softmax \\
  \hline
  \end{tabular}
\end{table}

Table~\ref{table:network} shows the architectures of the evaluation 
network and the policy network we use, which only differ in the first and last 
layers. The first layer of the policy network has an extra plane which
indicates which player's policy it is computing. The last layer of the 
evaluation network has two outputs, indicating if the state is a player 
1 or player 2 win, while the policy network has four outputs, each corresponding 
to one of four possible actions. The 
global averaging used after the convolutional layers does not use any extra weights, compared to a fully connected layer. The benefit is that the number of network parameters does not grow when the map size is increased. This allows for a network to be quickly pre-trained on 
smaller maps, and then fine-tuned on the larger target map.

\emph{Puppet Search} requires a forward model to examine the outcome 
of different actions and then choose the best one. Most RTS games do not have a dedicated forward 
model or simulator other than the game itself. This is usually too slow to be used 
in a search algorithm, or even unavailable due to technical constraints such as
closed source code or being tied to the graphics engine. Using a policy network
for script selection during game play allows us to bypass the need for a forward 
model of the game. Granted, the forward model is still required during the 
supervised training phase, but execution speed is less of an issue in this case,
because training is performed offline.
Training the network via reinforcement learning would remove this constraint 
completely.

Finally, with the policy network running significantly faster~(3ms 
versus a time budget of 100ms per frame for search-based agents) than 
\emph{Puppet Search} we can use the unused time to refine tactics. While the scripts 
used by \emph{Puppet Search} and the policy network represent different 
strategic choices, they all share very similar tactical behaviour. Their tactical ability 
 is weak in comparison to state-of-the-art search-based bots, as 
previous results~\cite{barriga2017game} suggest.

For this reason, the proposed algorithm combines an FCN for strategic decisions and 
an adversarial search algorithm for tactics. The strategic component handles 
macro-management: unit production, workers, and sending combat units towards the 
opponent. The tactical component handles micro-management during combat.

\algrenewcommand\algorithmicindent{1em}
\algdef{SE}[PROC]{Proc}{EndProc}%
   [5]{\parbox[t]{.65\linewidth}{\algorithmicprocedure\ \textproc{#1}}\ifthenelse{\equal{#2}{}}{}{\parbox[t]{.25\linewidth}{(#2\\#3\\#4\\#5)}}}%
   {\algorithmicend\ \algorithmicprocedure}%
\begin{algorithm}[b]
\caption{Combined Strategy and Tactics}\label{alg:stratTact}
\begin{algorithmic}[1]\raggedright
\Proc{\mbox{getCombinedAction}}{$state, stratAI,$}{$ tactAI,$}{$ stratTime,$}{$ tactTime$}

 \State $limState \gets$ \Call{extractCombat}{$state$}\label{l:trunc}

 \If{\Call{isEmpty}{$limState$}}
 	\State \Call{setTime}{$stratAI, stratTime + tactTime$}
 \Else	 
 	\State \Call{setTime}{$stratAI, stratTime$}
 	\State \Call{setTime}{$tactAI, tactTime$}
 \EndIf

 \State $stratActions \gets$ \Call{getAction}{$stratAI, state$}\label{l:strat}
 \State $tactActions \gets$ \Call{getAction}{$tactAI, limState$}\label{l:tact}
 
 \State \Return \Call{merge}{$stratActions, tactActions$}\label{l:merge} 
\EndProc
\end{algorithmic}
\end{algorithm}

The complete procedure is described by Algorithm~\ref{alg:stratTact}. It first
builds a limited view of the game state, which only includes units that are
close to opposing units~(line~\ref{l:trunc}). If this limited state is empty,
all available computation time is assigned to the strategic algorithm,
otherwise, both algorithms receive a fraction of the total time
available. This fraction is decided empirically for each particular algorithm
combination. Then, in line~\ref{l:strat} the strategic algorithm is used to
compute actions for all units in the state, followed by the tactical algorithm
that computes actions for units in the limited state. Finally, the actions are
merged~(line~\ref{l:merge}) by replacing the strategic action in case both
algorithms produced actions for a particular unit.

\section{Experiments and Results}\label{sec:results}

All experiments were performed in machines running Fedora 25, with an
Intel Core i7-7700K CPU, with 32GB of RAM and an NVIDIA GeForce
GTX 1070 with 8GB of RAM.
The Java version used for $\mu$RTS was OpenJDK 1.8.0, Caffe git commit 365ac88
was compiled with g++ 5.3.0, and pycaffe was run using python 2.7.13.

The \emph{Puppet Search} version we used for all the following experiments utilizes alpha-beta search over a single choice point with four options. The four options 
are \emph{WorkerRush}, \emph{LightRush}, \emph{RangedRush} and \emph{HeavyRush}, and were also used as baselines in the following experiments. More details about these scripts can be found in~\cite{stanescu2016evaluating}.

Two other recent algorithms were also used as benchmarks, Na\"ıveMCTS~\cite{ontanon2013naive} and Adversarial Hierarchical Task Networks (AHTNs)~\cite{ontanon2015adversarial}. Na\"ıveMCTS is an MCTS variant with a sampling strategy that exploits the tree structure of Combinatorial Multi-Armed Bandits --- bandit problems with multiple variables. Applied to RTS games, each variable represents a unit, and the legal actions for each of those units are the values that each variable can take. Na\"iveMCTS outperforms other game tree search algorithms on small scenarios. AHTNs are an alternative approach, similar to \emph{Puppet Search}, that instead of sampling from the full action space, uses scripted actions to reduce the search space. It combines minimax tree search with HTN planning.

All experiments were performed on 128x128 maps ported from the \emph{StarCraft: Brood War} maps used for the AIIDE competition. These maps, as well as implementations of \emph{Puppet Search}, the four scripts, AHTN and Na\"iveMCTS are readily available in the $\mu$RTS repository. 

\subsection{State Evaluation Network}
The data for training the evaluation network was generated by running 
games between a set of bots using 5 
different maps, each with 12 different starting positions. Ties were discarded, 
and the remaining games were split into 2190 training games, and 262 test games. 
12 game states were randomly sampled from each game, for a total of 26,280 
training samples and 3,144 test samples. Data is labelled by a Boolean value indicating whether the first player won. All evaluation functions were trained on the same 
dataset.

The network's weights are initialized using Xavier initialization~\cite{glorot2010understanding}. We used adaptive moment estimation (ADAM)~\cite{DBLP:journals/corr/KingmaB14} with default values of $\beta_1 = 0.9, \beta_2 = 0.999, \epsilon = 10^{-8}$ and a base learning rate of $10^{-4}$. The batch size was 256.

\begin{figure}[b!]
\centerline{\includegraphics[width=\columnwidth]{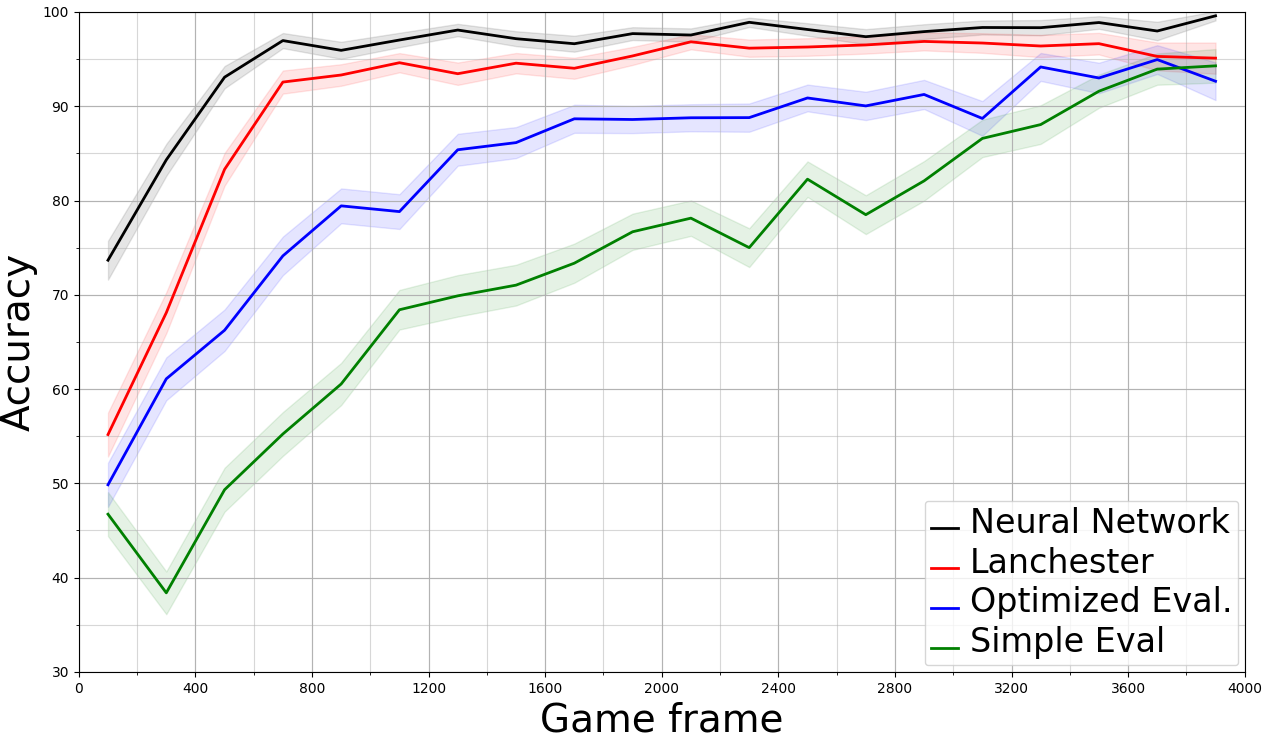}}
\caption{Comparison of evaluation accuracy between the neural network and the 
built-in evaluation function in $\mu$RTS.
The accuracy of predicting the game winner is plotted against game time.
Results are 
aggregated in 200-frame buckets. Shaded areas represent one
standard error.} \label{fig:accuracy}
\end{figure}

The evaluation network reaches 95\% accuracy in classifying samples as wins or losses. 
Figure~\ref{fig:accuracy} shows the accuracy of different evaluation functions 
as game time progresses. The functions compared are the evaluation network, 
Lanchester~\cite{stanescu2015}, the simple linear evaluation with hard-coded 
weights that comes with $\mu$RTS, and a version of the simple evaluation with 
weights optimized using logistic regression. The network's accuracy is even 
higher than previous results in 8x8 maps~\cite{stanescu2016evaluating}. The 
accuracy drop of the simple evaluation in the early game happens because it 
does not take into account units 
currently being built. If a player invests 
resources in units or buildings that take a long time to complete, its score 
lowers, despite the stronger resulting position after their completion. The 
other functions learn appropriate weights to mitigate this issue. 

Table~\ref{table:evalstime} shows the performance of \emph{PuppetSearch} when 
using the Lanchester evaluation function and the neural network. The
performance of the network is significantly better (P-value = 0.0011) than 
Lanchester's, even though the network is three orders of magnitude slower.
Evaluating a game state using Lanchester takes an average of 2.7$\mu$s, while 
the evaluation network uses 2,574$\mu$s.

\begin{table*}
\rowcolors{2}{lightgray}{white}
\begin{center}
{\caption{Policy network versus Puppet Search: round-robin tournament using 60 
    different starting positions per match-up.}\label{table:policy}}
\medskip
\begin{tabular}{|l|c|c|c|c|c|c||c|}
\hline
&PS&Policy&Light&Heavy&Ranged&Worker&Avg.\\
&&Net.&Rush&Rush&Rush&Rush&\\
\hline
PS	        &-	    &55.8	&87.5	&66.67   &91.7	&93.3	&65.8\\
Policy net.	&44.2	 &-	    &94.2	&71.7    &100    &61.7	&61.9\\
\hline
LightRush	&12.5	&5.8	&-	    &71.7    &100	&100	&48.3\\
HeavyRush	&33.3	&28.3	&28.3	&-       &100	&100	&48.3\\
RangedRush	&8.3	&0	    &0	    &0       &-	    &100	&18.1\\
WorkerRush	&6.7	&38.3	&0	    &0       &0	    &-	    &7.5\\
\hline
\end{tabular}
\end{center}

\rowcolors{2}{lightgray}{white}
\begin{center}
{\caption{Mixed Strategy/Tactics agents: round-robin tournament using 60 
    different starting positions per match-up.}\label{table:stratact}}
\medskip
\begin{tabular}{|l|c|c|c|c|c|c|c|c|c|c||c|}
\hline
&Policy&PS&PS&Policy&Light&Heavy&Ranged&AHTN&Worker&Na\"ive&Avg.\\
&Na\"ive&Na\"ive&&Network&Rush&Rush&Rush&P&Rush&MCTS&\\
\hline		
Policy net.-Na\"ive	&-		&56.7	&97.5	&100.0	&100.0	&95.8	&100.0	&72.5	&74.2	&98.3	&88.3\\
PS-Na\"ive		    &43.3	&-		&81.7	&79.2	&90.0	&94.2	&93.3	&90.0	&90.8	&93.3	&84.0\\
\hline
PS		            &2.5	&18.3	&-		&63.3	&86.7	&69.2	&92.5	&96.7	&95.0	&93.3	&68.6\\
Policy net.		    &0.0	&20.8	&36.7	&-		&94.2	&71.7	&100.0	&57.5	&61.7	&97.5	&60.0\\
\hline
LightRush		    &0.0	&10.0	&13.3	&5.8	&-		&71.7	&100.0	&100.0	&100.0	&96.7	&55.3\\
HeavyRush		    &4.2	&5.8	&30.8	&28.3	&28.3	&-		&100.0	&100.0	&100.0	&74.2	&52.4\\
RangedRush		    &0.0	&6.7	&7.5	&0.0	&0.0	&0.0	&-		&100.0	&100.0	&86.7	&33.4\\
AHTN-P		        &27.5	&10.0	&3.3	&42.5	&0.0	&0.0	&0.0	&-		&64.2	&68.3	&24.0\\
WorkerRush		    &25.8	&9.2	&5.0	&38.3	&0.0	&0.0	&0.0	&35.8	&-		&71.7	&20.6\\
Na\"iveMCTS		    &1.7	&6.7	&6.7	&2.5	&3.3	&25.8	&13.3	&31.7	&28.3	&-		&13.3\\
\hline
\end{tabular}
\end{center}
\vspace{-1em}
\end{table*}
Table~\ref{table:evalsfixed} shows the same comparison, but with \emph{Puppet 
Search} searching to a fixed depth of 4, rather than having 
100ms per frame. The advantage of the neural network is much more clear, 
as execution speed does not matter in this case. (P-value = 0.0044) 

\subsection{Policy Network}
We used the same procedure as in the previous subsection, but now we labelled 
the samples with the outcome of a 10 second \emph{Puppet Search} using the 
evaluation network. The resulting policy network has an accuracy 
for predicting the correct puppet move of 73\%, and a 95\% accuracy for 
predicting any of the top 2 moves.

Table~\ref{table:policy} shows the policy network coming close to 
\emph{Puppet Search} and defeating all the scripts.

\begin{table}[b!]
\rowcolors{2}{lightgray}{white}%
\begin{center}%
  \vspace{-0.65em}
\caption{Evaluation network versus Lanchester: round-robin tournament using  
60 different starting positions per match-up and
\textbf{100ms of computation time.}}\label{table:evalstime}
\medskip
\begin{tabular}{|l|c|c|c|c||c|}
\hline
&PS&PS&Light&Heavy&Avg.\\
&CNN&Lanc.&Rush&Rush&\\
\hline
PS CNN		            &-    &59.2	&89.2	&72.5	&73.6\\
PS Lanc.	           	&40.8 &-	&64.2	&67.5	&57.5\\
\hline
LightRush		       	&10.8 &35.8	&-	&71.7	&39.4\\
HeavyRush		       	&27.5 &32.5	&28.3	&-	&29.4\\
\hline
\end{tabular}
\end{center}

\rowcolors{2}{lightgray}{white}
\begin{center}
\caption{Evaluation network versus Lanchester: round-robin tournament on 
20 different starting positions per match-up, 
searching to \textbf{depth 4.}}\label{table:evalsfixed}
\medskip
\begin{tabular}{|l|c|c|c|c||c|}
\hline
&PS&PS&Light&Heavy&Avg.\\
&CNN&Lanc.&Rush&Rush&\\
\hline
PS CNN		    &-	    &80	    &95	    &82.5	&85.8\\
PS Lanc.		&20	    &-	    &82.5	&90	    &64.2\\
\hline
LightRush		&5	    &17.5	&-	    &70	    &30.8\\
HeavyRush		&17.5	&10	    &30	    &-	    &19.2\\
\hline
\end{tabular}
\end{center}
\end{table}

\subsection{Strategy and Tactics}\label{sec:strategyandtactics}

Finally, we compare the performance of the policy network and \emph{Puppet
  Search} as the strategic part of a combined strategic/tactical agent. We
will do so by assigning a fraction of the allotted time to the strategic algorithm
and the remainder to the tactical algorithm, which will be Na\"iveMCTS in our
experiments. We expect the policy network to perform better in this scenario,
as it runs significantly faster than \emph{Puppet Search} while
maintaining similar action performance.

The best time split between strategic and tactical algorithm was determined
experimentally to be 20\% for \emph{Puppet Search} and 80\% for Na\"iveMCTS. 
The policy network uses a fixed time (around 3ms), and the
remaining time is assigned to the tactical search.

Table~\ref{table:stratact} shows that both strategic algorithms greatly
benefit from blending with a tactical algorithm. The gains are more
substantial for the policy network, which now scores 56.7\% against its \emph{Puppet
  Search} counterpart. It also has a 4.3\% higher
overall win rate despite markedly poorer results against WorkerRush and
AHTN-P. These seems to be due to a strategic mistake on the part of the
policy network, which, if its cause can be detected and corrected, would
lead to even higher performance.

\section{Conclusions and Future Work}\label{sec:concl}
We have extended previous research that used CNNs to accurately evaluate RTS
game states in small maps to larger map sizes usually used in commercial RTS
games. The average win prediction accuracy at all game times is higher
compared to smaller scenarios. This is probably the case because strategic
decisions are more important than tactical decisions in larger maps, and
strategic development is easier to quantify by the network. Although the
network is several orders of magnitude slower than competing simpler
evaluation functions, its accuracy makes it more effective. When the
\emph{Puppet Search} high-level adversarial search algorithm uses the CNN, its
performance is better than when using simpler but faster functions.

We also trained a policy network to predict the outcome of \emph{Puppet
  Search}. The win rate of the resulting network is similar to that of the
original search, with some exceptions against specific opponents. However,
while slightly weaker in playing strength, a feed-forward network pass is much
faster. This speed increase created the opportunity for using the saved time
to fix the shortcomings introduced by high-level abstractions. A tactical
search algorithm can micro-manage units in contact with the enemy, while the
policy chosen by the network handles routine tasks~(mining, marching units
toward the opponent) and strategic tasks~(training new units). The resulting
agent was shown to be stronger than the policy network alone in all tested
scenarios, but can only partially compensate for the network's weaknesses
against specific opponents.

Looking into the future, we recognize that most tactical search algorithms,
like the MCTS variant we used, have the drawback of requiring a forward model
of the game. Using machine learning techniques to make tactical decisions
would eliminate this requirement. However, this has proved to be a difficult
goal, as previous attempts by other researchers have had limited success on
simple scenarios~\cite{usunier2016episodic,synnaeve2016multiscale}.  Recent
research avenues based on integrating concepts such as
communication~\cite{Sukhbaatar2016}, unit grouping and bidirectional recurrent
neural networks~\cite{Peng2017} suggest that strong tactical networks might
soon be available.

The network architecture presented in this paper, being fully convolutional,
can be used on maps of any (reasonable) size without increasing its number of
parameters. Hence, future research could include assessing the speed-up
obtained by taking advantage of \emph{transfer learning} from smaller maps to
larger ones. Also of interest would be to determine whether different map
sizes can be mixed within a training set. It would also be interesting to
investigate the performance of the networks on maps that have not previously
been seen during training .

Because the policy network exhibits some weaknesses against specific
opponents, further experiments should be performed to establish whether this
is due to a lack of appropriate game state samples in the training data or
other reasons. A related issue is our reliance on labelled training data, which
could be resolved by using reinforcement learning techniques, such as
DQN~(deep Q network) learning. However, full RTS games are difficult for these
techniques, mainly because the only available reward is the outcome of the
game. In addition, action choices near the endgame (close to the reward), have
very little impact on the outcome of the game, while early ones (when there is
no reward), matter most. There are several strategies available that could
help overcome these issues, such as curriculum
learning~\cite{bengio2009curriculum}, reward
shaping~\cite{devlin2011empirical}, or implementing double
DQN learning~\cite{hasselt2016deep}. These strategies have proved useful on adversarial
games, games with sparse rewards, or temporally extended planning problems
respectively.

\bibliographystyle{aaai} 
\bibliography{refs}

\vfill
\end{document}